\newcommand{\abstractcommand}{%
In this paper we derive an explicit formula for calculating the marginal
likelihood of a given factorization of a categorical dataset. Since the
marginal likelihood is proportional to the posterior probability of the
factorization, these likelihoods can be used to order all possible
factorizations and select the ``best'' way to factor the distribution from
which the dataset is drawn.  The best factorization can then be used to
construct a Bayes classifier which benefits from factoring out mutually
independent sets of variables.}

\ifdefined\ba
\documentclass[ba]{imsart}

\pubyear{2021}
\volume{TBA}
\issue{TBA}
\firstpage{1}
\lastpage{1}

\usepackage{amsthm}
\usepackage{amsmath}
\usepackage{natbib}
\usepackage[colorlinks,citecolor=blue,urlcolor=blue,filecolor=blue,backref=page]{hyperref}
\usepackage{graphicx}
\usepackage{bm}

\startlocaldefs
\numberwithin{equation}{section}
\theoremstyle{plain}

\endlocaldefs

\begin{document}

\begin{frontmatter}
\title{Factoring Multidimensional Data to Create a Sophisticated Bayes Classifier}
\runtitle{Factoring Multidimensional Data}

\begin{aug}
\author{\fnms{Anthony} \snm{LaTorre}\thanksref{addr1,addr2}\ead[label=e1]{tlatorre9 at gmail dot com}}

\runauthor{A. LaTorre}

\address[addr1]{Department of Physics, University of Chicago, Chicago IL 60637
    \printead{e1} % print email address of "e1"
}

\address[addr2]{Division of Math, Physics, and Astronomy, California Institute of Technology, Pasadena, CA 91125
}

\end{aug}

\begin{abstract}
\abstractcommand
\end{abstract}

\begin{keyword}
\kwd{multidimensional}
\kwd{factor}
\kwd{bayesian}
\kwd{classifier}
\end{keyword}

\end{frontmatter}
\else
\documentclass[12pt]{article}
\usepackage{fullpage}
\usepackage{amsmath}
\usepackage{bm}
\usepackage{authblk}
\title{Factoring Multidimensional Data to Create a Sophisticated Bayes Classifier}
\date{\today}
\author[1,2]{Anthony LaTorre\thanks{Electronic address: \texttt{tlatorre9 at gmail dot com}}}
\affil[1]{Department of Physics, University of Chicago, Chicago IL 60637}
\affil[2]{Division of Math, Physics, and Astronomy, California Institute of Technology, Pasadena, CA 91125}
\renewcommand\Affilfont{\itshape\small}
\begin{document}
\maketitle
\begin{abstract}
\abstractcommand
\end{abstract}

\fi

\section{Introduction}
We wish to investigate the most efficient way to compute the probability of
some variable conditioned on other variables, i.e. to calculate $P(y\mid
\bm{x})$. This can be done using the definition of conditional probability
which says:
\begin{equation}
P(y\mid\bm{x}) = \frac{P(y,\bm{x})}{P(\bm{x})}.
\label{eq:bayes}
\end{equation}
The denominator in Equation~\ref{eq:bayes} is a scaling factor which doesn't
depend on $y$, so our focus will be the estimation of $P(y,\bm{x})$, which can
also be written as
\begin{equation}
P(y,\bm{x}) = P(y)P(\bm{x}\mid y).
\end{equation}
Since we assume $y$ is a single variable, we can efficiently estimate $P(y)$,
so the problem can also be reduced to estimating $P(\bm{x}\mid y)$.

The simplest approach to computing $P(y,\bm{x})$ would be to construct an $N$
dimensional histogram, assuming we have $N-1$ random variables in $\bm{x}$, and
computing the relative frequency of events ending up in the bin $(y,\bm{x})$.
This method has the advantage that we don't have to make any assumptions about
the distribution from which the data is drawn. The drawback is that the error
on the prediction of $y$ will scale exponentially like $K^{N/2}$, where $K$ is
the number of categories or bins for each variable. This makes intuitive sense
since for a dataset with a fixed number of samples, the number of bins grows
exponentially with the number of variables, and therefore we should expect a
large error because the statistics in each bin will be low.

One common solution to this problem is to use a Naive Bayes
classifier~\cite{idiot} where all the $X_i$ variables are assumed to be
mutually independent conditional on $y$. With this assumption we can rewrite
Equation~\ref{eq:bayes} as
\begin{equation}
P(y\mid\bm{x}) = \frac{1}{P(\bm{x})} P(y) \prod_i P(x_i\mid y).
\end{equation}
The error on the prediction in this case only grows like $\sqrt{2N}$. This also
makes intuitive sense since we have reduced the calculation from looking up the
probability in a $N$ dimensional histogram to a product of $N$ 1-dimensional
histograms. However, the accuracy of this estimate will depend on how good the
independent approximation is.

In this paper, we propose a solution to this problem by asking how the
distribution $P(\bm{x}\mid y)$ can be factored. For example, supposing we had
just two variables $X_1$ and $X_2$, we can ask whether it can be factored as
$P(x_1\mid y)P(x_2\mid y)$ or whether we need to consider the full probability
$P(x_1,x_2\mid y)$. To distinguish between these two models, we consider
ordering them by the posterior probability of the factorization~\cite{jaynes}.
For the example just described this involves calculating two probabilities:
$P(M_I\mid D)$ and $P(M_D\mid D)$, where $M_I$ stands for the independent model
which can be factored and $M_D$ stands for the dependent model. In particular,
we show that for a categorical distribution, there is a simple analytical
solution for computing the marginal likelihood\footnote{By marginal likelihood
we mean the likelihood of the model independent of the specific parameters
governing the model.} of a given model, i.e. $P(D\mid M)$ which is proportional
to the posterior probability $P(M\mid D)$ up to a constant scaling factor and
some prior over the models.

For the general case of $N$ variables, we have to consider all possible ways to
factor $P(\bm{x}\mid y)$. This is equivalent to considering all possible
groupings of the variables, where within each group the variables are
considered to be dependent, but the different groups are treated as
independent. For example, all possible groupings of 4 variables are shown in
Table~\ref{table:groupings}.

\begin{table}
\centering
\begin{tabular}{c}
$(X_1,X_2,X_3,X_4)$ \\
$(X_1),(X_2,X_3,X_4)$ \\
$(X_2),(X_1,X_3,X_4)$ \\
$(X_3),(X_1,X_2,X_4)$ \\
$(X_4),(X_1,X_2,X_3)$ \\
$(X_1,X_2),(X_3,X_4)$ \\
$(X_1,X_3),(X_2,X_4)$ \\
$(X_1,X_4),(X_2,X_3)$ \\
$(X_1),(X_2),(X_3,X_4)$ \\
$(X_1),(X_3),(X_2,X_4)$ \\
$(X_1),(X_4),(X_2,X_3)$ \\
$(X_2),(X_3),(X_1,X_4)$ \\
$(X_2),(X_4),(X_1,X_3)$ \\
$(X_3),(X_4),(X_1,X_2)$ \\
$(X_1),(X_2),(X_3),(X_4)$ \\
\end{tabular}
\caption{All possible groupings of 4 variables. Each group represents a different way to factor the probability distribution $P(x_1,x_2,x_3,x_4)$.}
\label{table:groupings}
\end{table}

%Therefore, in this paper we propose an analytical solution for asking which
%variables are independent from which other variables through the use of Bayes
%Factors. More specifically, show how to calculate the marginal likelihood of a
%dataset for the hypothesis of specific grouping of which variables are
%correlated, which can be used to construct Bayes Factors for different
%groupings and thereby determine which variables are independent from one
%another.

%In Section~\ref{sec:bayes} we give an introduction to Bayes Factors and how
%they can be used to select the best model. In Section~\ref{sec:calculation} we
%show how to compute the key term $P(D|M)$ necessary to compute the Bayes factor
%for different models. In Section~\ref{sec:two-binary-variables} we explicitly
%calculate the Bayes factor for two binary random variables, and in
%Section~\ref{sec:conclusion} we give concluding remarks.

The rest of the paper is organized as follows: in Section~\ref{sec:calculation}
we consider the general problem of how to factor a distribution $P(\bm{x})$ by
computing the marginal likelihood $P(D\mid M)$ for a given factorization. In
Section~\ref{sec:two-binary-variables} we explicitly calculate these terms for
two binary random variables and then give a formula for the Bayes factor to
distinguish between them. Finally, in Section~\ref{sec:conclusion} we give
concluding remarks.

\section{Calculation}
\label{sec:calculation}
In this section, we consider the general problem of factoring a distribution of
$N$ categorical random variables labelled $(X_0,X_1,\ldots,X_N)$. To do so, we
choose the ``best'' factorization where the ordering of the different
factorizations is determined by the posterior probability of a given
factorization. If we label a given factorization as a model $M$, we wish to
compute $P(M\mid D)$ where $D$ stands for the data. This posterior is related to
the marginal likelihood $P(D\mid M)$ as follows:
\begin{equation}
P(M\mid D) = \frac{P(D\mid M)P(M)}{P(D)}.
\label{eq:posterior}
\end{equation}
The first term in the numerator of Equation~\ref{eq:posterior} is the marginal
likelihood, the second term in the numerator represents any prior we may have,
and the denominator is simply a scaling factor which is not relevant if we only
consider ordering the models.  So, to order the models we need to compute the
marginal likelihood $P(D\mid M)$ and then we can multiply it by any prior we
might have (or simply use the likelihood directly if we have no preference for
any particular factorization).

We will represent each possible factorization as a particular grouping of the
$N$ variables. For example, the possible groupings for 4 random variables are
shown in Table~\ref{table:groupings}. For a given factorization $M$, we will
label each group within the factorization as $g_i$, where $i$ runs over all the
different groups. For example, one possible factorization of 4 random variables
is $((X_1,X_2),(X_3),(X_4))$. For this case, the group $(X_1,X_2)$ will be
referred to as $g_0$, the next group $(X_3)$ as $g_1$, and $(X_4)$ as $g_2$.

The probability distribution for each group $g_i$, can be specified by the
variables $p_{i,j}$ labeling the probability of landing in the $j^\mathrm{th}$
bin for group $g_i$. When referring to these probabilities for a general group
$g$ we will also sometimes denote these probabilities as $p_{g,j}$. The number
of bins in each group is equal to
\[
\eta_i = \prod_{X\in g_i} |X|.
\]
where $|X|$ is the number of possible values for the random variable $X$. For
the previous example, assuming each variable can take on only two discrete
values, there would be 4 probabilities for the first group,
$p_{0,0},p_{0,1},p_{0,2},p_{0,3}$, two probabilities for the second group
$p_{1,0},p_{1,1}$, and two probabilities for the last group $p_{2,0},p_{2,1}$.
We will sometimes denote these values as vectors, i.e. $\bm{p_0}$, $\bm{p_1}$,
and $\bm{p_2}$. When referring to a general group $g$, we will simply write
$\bm{p_g}$.

We begin by considering the marginal likelihood and expanding it in terms of
the probabilities $p_{i,j}$
\begin{align}
P(D\mid M) &= \int_{p_{0,1}}\int_{p_{0,2}}\cdots\int_{p_{m,\eta_m}}P(D\mid p_{i,j},M)P(p_{i,j}\mid M),
\label{eq:marginal-likelihood}
\end{align}
where $m$ is equal to the number of groups. The two terms on the right are the
full likelihood and the prior for the model parameters $p_{i,j}$ for the
factorization $M$.

For the second term in Equation~\ref{eq:marginal-likelihood}, we assume a
flat\footnote{The Dirichlet distribution with $\bm{\alpha} = \bm{1}$ is
completely flat and so is the only non-informative prior where our only
constraint is that the sum of the probabilities must add to 1. It also happens
to be a good choice since it is the conjugate prior of the multinomial
distribution.} Dirichlet prior for the probabilities within each group,
\begin{equation}
P(p_{i,j}\mid M) = \prod_{g\in M} \mathrm{Dir}(\bm{p_g},\bm{1}).
\label{eq:prior}
\end{equation}
The Dirichlet distribution is given by the equation:
\begin{equation}
\mathrm{Dir}(\bm{\alpha},\bm{x}) = \frac{1}{B(\bm{\alpha})}\prod_i x_i^{\alpha_i-1}
\end{equation}
where $B$ is the multivariate Beta function given by
\begin{align}
B(\bm{\alpha}) &= \frac{\prod_i \Gamma(\alpha_i)}{\Gamma(\sum_i \alpha_i)}.
\end{align}

The first term in Equation~\ref{eq:marginal-likelihood} is the full likelihood
of observing a dataset $D$, which is given by the multinomial distribution
\begin{equation}
P(D\mid p_{i,j},M) = \frac{N!}{\prod_i n_i!} \prod_i p_i^{n_i}.
\label{eq:p-group}
\end{equation}
Note that in this last expression, the $i$ variable runs over all possible
combinations of the values for the $\bm{X}$ variables, $n_i$ represents the
number of samples with outcome $i$, and $p_i$ represents the probability of
getting the outcome $i$. Table~\ref{table:p} shows all the terms in this
equation for a factorization of 4 random variables as $(X_1,X_2),(X_3),(X_4)$.
Next, we rewrite Equation~\ref{eq:p-group} grouping the $p$ variables for
groups together
\begin{equation}
P(D\mid p_{i,j},M) = \frac{N!}{\prod_i n_i!} \prod_{g\in G} \prod_i p_{g,i}^{n_{g,i}}
\label{eq:n-group}
\end{equation}
where the $n_{g,i}$ term is equal to the number of data values which have the
$i^{\mathrm{th}}$ combination of values for the group $g$ marginalized over all the
other variables. This is equivalent to the $i^{\mathrm{th}}$ bin in the flattened
histogram of the dataset when only looking at the variables in the group $g$.

\begin{table}
\centering
\begin{tabular}{cccc}
i  & $\bm{x}$ & $p_i \rightarrow p_{i,j}$ \\ \hline
0  & 00~0~0     & $p_0     = p_{0,0} \cdot p_{1,0} \cdot p_{2,0}$ \\
1  & 00~0~1     & $p_1     = p_{0,0} \cdot p_{1,0} \cdot p_{2,1}$ \\
2  & 00~1~0     & $p_2     = p_{0,0} \cdot p_{1,1} \cdot p_{2,0}$ \\
3  & 00~1~1     & $p_3     = p_{0,0} \cdot p_{1,1} \cdot p_{2,1}$ \\
4  & 01~0~0     & $p_4     = p_{0,1} \cdot p_{1,0} \cdot p_{2,0}$ \\
5  & 01~0~1     & $p_5     = p_{0,1} \cdot p_{1,0} \cdot p_{2,1}$ \\
6  & 01~1~0     & $p_6     = p_{0,1} \cdot p_{1,1} \cdot p_{2,0}$ \\
7  & 01~1~1     & $p_7     = p_{0,1} \cdot p_{1,1} \cdot p_{2,1}$ \\
8  & 10~0~0     & $p_8     = p_{0,2} \cdot p_{1,0} \cdot p_{2,0}$ \\
9  & 10~0~1     & $p_9     = p_{0,2} \cdot p_{1,0} \cdot p_{2,1}$ \\
10 & 10~1~0     & $p_{10}  = p_{0,2} \cdot p_{1,1} \cdot p_{2,0}$ \\
11 & 10~1~1     & $p_{11}  = p_{0,2} \cdot p_{1,1} \cdot p_{2,1}$ \\
12 & 11~0~0     & $p_{12}  = p_{0,3} \cdot p_{1,0} \cdot p_{2,0}$ \\
13 & 11~0~1     & $p_{13}  = p_{0,3} \cdot p_{1,0} \cdot p_{2,1}$ \\
14 & 11~1~0     & $p_{14}  = p_{0,3} \cdot p_{1,1} \cdot p_{2,0}$ \\
15 & 11~1~1     & $p_{15}  = p_{0,3} \cdot p_{1,1} \cdot p_{2,1}$ \\
\end{tabular}
\caption{Table showing the 16 terms in the likelihood expression for four binary random variables. The second column shows the different possible values for the random variables $X_1$, $X_2$, $X_3$, and $X_4$ as a binary string. The third column shows the values for the probability written as they appear in Equation~\ref{eq:p-group} and then expanded in terms of the probabilities for the factorization $(X_1,X_2),(X_3),(X_4)$ in Equation~\ref{eq:n-group}.}
\label{table:p}
\end{table}

Plugging in the prior (Equation~\ref{eq:prior}) and likelihood
(Equation~\ref{eq:n-group}) into Equation~\ref{eq:marginal-likelihood} we get
\begin{align}
P(D\mid M) &= \int_{p_{0,1}}\int_{p_{0,2}}\cdots\int_{p_{m,\eta_m}} \frac{N!}{\prod_i n_i!} \prod_{g\in M} \mathrm{Dir}(\bm{p_g},\bm{1}) \prod_i p_{g,i}^{n_{g,i}}.
\label{posterior}
\end{align}
The last product in this expression can be written as
\begin{equation}
\mathrm{Dir}(\bm{p_g},\bm{1}) \prod_i p_{g,i}^{n_i} = \frac{1}{B(\bm{1})} \prod_i p_{g,i}^{n_i} = \mathrm{Dir}(\bm{p_g},\bm{1}+\bm{n})\frac{\mathrm{B}(\bm{1} + \bm{n})}{\mathrm{B}(\bm{1})}.
\end{equation}
Plugging this back into Equation~\ref{posterior} we get
\begin{align}
P(D\mid M) &= \int_{p_{0,1}}\int_{p_{0,2}}\cdots\int_{p_{m,\eta_m}}\frac{N!}{\prod_i n_i!} \prod_{g\in M} \mathrm{Dir}(\bm{p_g},\bm{1}+\bm{n})\frac{\mathrm{B}(\bm{1} + \bm{n_g})}{\mathrm{B}(\bm{1})}
\end{align}
which simplifies to 
\begin{align}
P(D\mid M) &= \frac{N!}{\prod_i n_i!} \prod_{g\in M}\frac{\mathrm{B}(\bm{1} + \bm{n_g})}{\mathrm{B}(\bm{1})}.
\label{eq:marginal-likelihood2}
\end{align}
Note that the length of the vector $\bm{1}$ in the denominator of
Equation~\ref{eq:marginal-likelihood2} is equal to $\eta_i$ for group $i$, i.e.
it has the same length as $\bm{p_g}$ and $\bm{n_g}$.

Equation~\ref{eq:marginal-likelihood2} is the main result of this paper which
gives an explicit formula for the calculation of the marginal likelihood of a
given factorization. There exist efficient routines for calculating the
multivariate Beta function, so calculating the marginal likelihood for a given
factorization is straightforward. In practice, it is useful to compute the log
of Equation~\ref{eq:marginal-likelihood2} when comparing factorizations since
otherwise the factorials can produce numbers which do not fit into standard 32
or 64 bit floats and integers.

%\begin{equation}
%n_{g,i} = \sum_{j=0}^{2^n - |g| - b_{g+1} - 1} \sum_{k=0}^{2^{b_g} - 1} n_{2^{b_g} \cdot j + k + i \cdot 2^{b_{g+1}}}.
%\end{equation}
%where $b_g$ is equal to the position of the start of the group in the ordered
%list. For example, for the grouping $(1,2),3,4$ the $b$ value for the first
%group is 4, for the second 2, and for the third 1. We define the $b_{g+1}$ for
%the last group to be 0.

\section{Two Binary Variables}
\label{sec:two-binary-variables}
\begin{table}
\centering
\begin{tabular}{ccc}
$x_1$ & $x_2$ & $n_i$ \\ \hline
0     & 0     & $n_1$ \\
0     & 1     & $n_2$ \\
1     & 0     & $n_3$ \\
1     & 1     & $n_4$
\end{tabular}
\caption{Table showing the mapping between the values for the random variables $\bm{x}$ and the numbers in Equation~\ref{eq:two-binary-variables-marginal-likelihood}.}.
\label{table:two-binary-variables}
\end{table}

In the simple case of 2 binary random variables, we get the result
\begin{align}
P(D\mid M_I) &= \frac{(n_1+n_2)!(n_1+n_3)!(n_2+n_4)!(n_3+n_4)!}{n_1!n_2!n_3!n_4!(N+1)(N+1)!}
\label{eq:two-binary-variables-marginal-likelihood} \\
P(D\mid M_D) &= \frac{6}{(N+3)(N+2)(N+1)}
\end{align}
where $M_I$ stands for the independent hypothesis, and $M_D$ stands for the
dependent hypothesis, the $n_i$ are the numbers of samples with the values
shown in Table~\ref{table:two-binary-variables}, and $N$ is the total number of
samples.

We can use these values to compute the Bayes Factor
\begin{align*}
K = \frac{P(D\mid M_I)}{P(D\mid M_D)} = \frac{(n_1+n_2)!(n_1+n_3)!(n_2+n_4)!(n_3+n_4)!(N+3)(N+2)}{n_1!n_2!n_3!n_4!(N+1)!6}.
\end{align*}
For values of $K$ less than 1 the data favor the correlated hypothesis, and for
values greater than 1 it favors the independent hypothesis.

\section{Conclusion}
\label{sec:conclusion}
We have given an explicit formula for calculating the marginal likelihood of a
given factorization of a dataset of categorical random variables. This formula
can be used to choose the best way to factor the conditional probability
$P(\bm{x}\mid y)$ and thus gives us a way to construct a Bayes classifier which
takes full advantage of mutually independent sets of variables.

Although it is possible in principle sort through all possible factorization
using Equation~\ref{eq:marginal-likelihood2}, the number of calculations of the
multivariate Beta function grows like the size of the powerset of the number of
variables. This means that in practice it is prohibitive to sort through all
factorizations for more than approximately 30 variables (where one would need
to calculate over a billion terms). However, we suspect that for real world
data there are ways to considerably reduce the number of factorizations to
check by using heuristics.

One particular area where we hope this technique will be of use is in high
energy physics analyses. For these analyses it is very common to define a broad
range of cuts designed to remove background events and keep events of interest.
Although multiple cuts are used, in the end the ultimate goal is often to
define a total cut efficiency and sacrifice, i.e. to determine what percentage
of background events are successfully cut and what fraction of signal events
are ``accidentally'' cut. If you have a full Monte Carlo simulation of both
background and signal it can be relatively easy to do this, but often some of
the background events cannot be simulated due to complexity or time
constraints. Therefore, it is often necessary to use a combined fit to a
sideband and the region of interest to fit for the cut efficiencies. The number
of terms in this fit will grow exponentially as the number of cuts increases,
and therefore this technique can be used to construct a model of for the cut
efficiencies with considerably fewer parameters.

\ifdefined\ba
\bibliographystyle{ba}
\bibliography{bayes}
\begin{acks}[Acknowledgments]
Thanks to Kevin Labe and Marie Vidal for multiple reviews of this paper.
\end{acks}
\else
\section*{Acknowledgments}
Thanks to Kevin Labe and Marie Vidal for multiple reviews of this paper.
\bibliographystyle{plain}
\bibliography{bayes}
\fi

\end{document}